\documentclass[11pt]{article}
\usepackage[final]{acl}
\usepackage{times}
\usepackage{latexsym}
\usepackage{amsmath}
\usepackage{amssymb}
\usepackage{graphicx}
\usepackage{tikz}
\usepackage{booktabs}
\usepackage{multirow}
\usepackage{xcolor}
\usepackage{subcaption}
\usepackage{enumitem}
\usepackage{url}
\usepackage{array}
\usepackage{float}
\usepackage{listings}
\usepackage{multicol}
\lstdefinestyle{promptbox}{
  basicstyle=\footnotesize\ttfamily,
  breaklines=true,
  columns=fullflexible,
  keepspaces=true,
  showstringspaces=false,
  frame=single,
  framesep=5pt,
  xleftmargin=6pt,
  xrightmargin=3pt,
  aboveskip=6pt,
  belowskip=2pt,
  postbreak=\mbox{\textcolor{gray}{$\hookrightarrow$}\space},
}
\newcommand{\system}{\textsc{CogniLife}}

\title{\system: Continuous Cognitive Distillation for Deployed Multi-Expert Agent Marketplaces}

\author{\normalfont\small
Xuwei Tan, Jingchun Huang, Ming Cai, Shuangquan Lyu, Tiange Wang,\\
\small Yinzhi Wang, Ronghua Shi, Jingwei Song, Yinghui Xia, Hehua Fan,\\
\small Yifan Zhou, Hanxin Chen, JIWEI SONG, Yang Jingsong\textsuperscript{\textdagger}, TIANYU SHI\textsuperscript{\textdagger}\\
\small \textsuperscript{\textdagger}Corresponding authors: Tianyu Shi (\texttt{tianyu.shi3@mcgill.ca}); Yang Jingsong (\texttt{edward.yang@agentspro.cn})
}
\begin{document}
\maketitle
\raggedbottom
\begin{abstract}
Multi-expert agent marketplaces---where creators publish domain-specific AI experts for end users to consult---are an emerging deployment pattern for language technology. A published expert can receive many requests yet little evidence about its creator, who remains active elsewhere: the \emph{cold-expert problem}. We frame continuous cognitive distillation as converting attributed creator observations into revisable decision rules, communication patterns, and domain procedures. \system{} proposes a cross-expert attribution design for this problem, with Creator Profiles for transferable creator patterns and Expert Journals for listing-specific knowledge; this design is not yet implemented or evaluated. A deployed platform partially realizes the surrounding lifecycle through creator-authorized authoring, capability-owned proposal filtering, and reversible consolidation of eligible owner-scoped memory. On \textsc{ProfyBench-v1}, an 85-task offline benchmark built from replay seeds and controlled internal tests, the Profy Full configuration as a whole achieves a mean overall score of 4.65/5 from the primary LLM judge, compared with 4.01 for Raw LLM and 4.28 for Expert Prompt, with the largest gains in completeness, specificity, and missing-information tasks. These configuration-level scores do not test cold-expert mitigation, cross-expert transfer, or longitudinal distillation.
\end{abstract}

\section{Introduction}
\label{sec:intro}

\emph{Multi-expert agent marketplaces} are emerging on AI platforms that enable creators to publish and monetize their own agents.\footnote{Examples include the Poe Creator Platform (\url{https://creator.poe.com/docs/intro}) and Delphi's Digital Minds (\url{https://www.delphi.co/blog/launch-your-digital-mind}).} Published experts should preserve their creators' behavioral commitments while serving different callers. Persona-conditioned dialogue \cite{zhang2018personalizing} uses textual profiles, while personalization work models heterogeneous user preferences \cite{zollo2025personalllm} or conditions outputs on user profiles \cite{salemi2024lamp}. Maintaining these services requires deciding whose behavior should refine which representation.

Consider a creator who publishes a translation expert but mostly consults coding and writing experts. The target can receive many requests yet remain sparse in relevant creator evidence: the \textbf{cold-expert problem}. Repeated creator corrections of ambiguity may transfer to translation, while coding-specific instructions stay local and other callers' preferences must not be mistaken for creator traits.

We define \emph{continuous cognitive distillation} as repeatedly abstracting attributed creator observations into persistent, revisable decision rules, communication patterns, and domain procedures. Our framework organizes evidence around the represented creator and scopes guidance to each expert through \emph{Seeding}, \emph{Immersion}, and \emph{Crystallization}.

\begin{figure*}[!t]
\centering
\includegraphics[width=\textwidth]{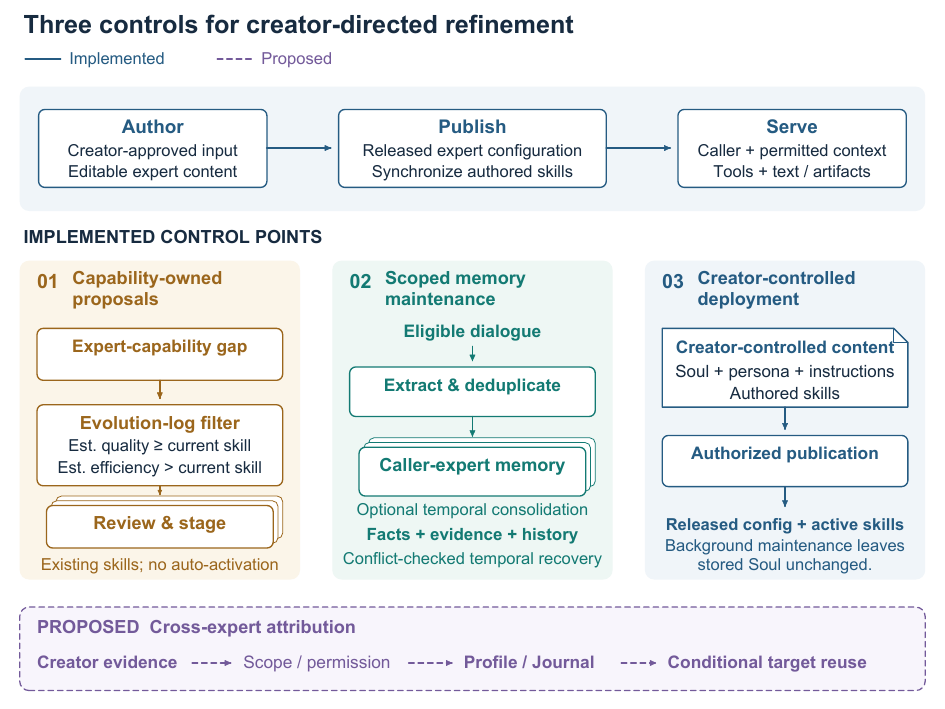}
\caption{Profy marketplace workflow. Solid panels show implemented skill proposals, scoped memory, and creator-authorized publication; dashed purple paths show proposed routing to Creator Profiles and Expert Journals. The routing is neither implemented nor evaluated. Skill activation and rollback have separate eligibility checks (Section~\ref{sec:maturity}; Appendix~\ref{app:implementation}).}
\label{fig:architecture}
\end{figure*}

Figure~\ref{fig:architecture} separates implemented platform controls from proposed cross-expert routing; Table~\ref{tab:deployment-challenges} maps deployment challenges to design responses.

Our contributions are threefold:
\begin{itemize}[leftmargin=*,nosep]
\item \textbf{Cold-expert formulation.} We define sparse creator evidence despite activity elsewhere and distinguish the represented creator, source interaction, target expert, and caller; mitigation remains untested.
\item \textbf{Creator-controlled maintenance.} We describe implemented capability-owned proposal filtering, eligible reversible owner-scoped memory maintenance, and staged revisions of existing skills; the cross-expert attribution protocol remains proposed and unevaluated.
\item \textbf{Expert-service evaluation.} We introduce \textsc{ProfyBench-v1}, comparing a Full configuration of supplied expert context and workflows with two baselines on 85 offline tasks, with the largest judged gains in completeness and specificity. We additionally report a supplementary 52-task customs-extraction benchmark with three runs per task, accompanied by an auditable example specifying source documents, structured outputs, and a frozen field-scoring contract.
\end{itemize}

\begin{table}[t]
\centering
\small
\begin{tabular}{@{}>{\raggedright\arraybackslash}p{0.27\linewidth}p{0.63\linewidth}@{}}
\toprule
\textbf{Challenge} (\textit{status}) & \textbf{\system{} design response} \\
\midrule
Tacit expertise \newline (\textit{partial}) & Elicit commitments, then derive revisable guidance from attributed evidence. \\
Cold experts \newline (\textit{partial}) & Seed sparse listings; propose reuse of relevant creator evidence. \\
Evidence attribution \newline (\textit{proposed}) & Resolve source, applicability, and permitted reuse before routing claims. \\
Persistent refinement \newline (\textit{partial}) & Consolidate scoped memory; evaluate and version skill updates. \\
Creator control \newline (\textit{implemented}) & Keep authored commitments separate from learned candidates and releases. \\
\bottomrule
\end{tabular}
\caption{Deployment challenges motivating \system{} and implementation status (partial: some steps implemented, others proposed or incomplete). \textsc{ProfyBench-v1} tests the Full configuration, not individual components.}
\label{tab:deployment-challenges}
\end{table}

\section{Related Work}

\paragraph{Personas and creator authoring.} PersonaChat conditions dialogue on textual profiles \cite{zhang2018personalizing}; trainable characters and role-specific tuning extend these representations \cite{shao2023character,wang2024rolellm}. PersonalLLM benchmarks adaptation to heterogeneous simulated user preferences \cite{zollo2025personalllm}, and LaMP personalizes outputs using user-profile records \cite{salemi2024lamp}. CloneBuilder supports creator configuration, material uploads, testing, and feedback \cite{lee2026selfclones}. Knowledge elicitation spans multiple techniques \cite{cooke1994varieties}; our design combines creator input with interaction evidence. Tacit expertise may resist verbalization \cite{polanyi1966tacit}, and reported attitudes need not predict situated action \cite{jerolmack2014talk}. Our lifecycle connects the represented creator's authored commitments to revisable guidance from later interactions.

\paragraph{Continual improvement and memory.} Lifelong dialogue research motivates continual adaptation \cite{liu2021lifelong}; Agent-in-the-Loop improves customer-support models through operational feedback \cite{zhao2025agent}. ACE incrementally revises context playbooks, while Mem0 extracts and maintains conversational memory \cite{zhang2026ace,chhikara2025mem0}. PersonaTree links interaction evidence, recurring patterns, and durable person-level claims through lifecycle memory operations \cite{hou2026personatree}. \system{} adopts similar updating and evidence-abstraction operations while emphasizing ownership, intervention, and recovery controls.

\paragraph{Sharing and attribution.} ShareMem shares reusable experience while grounding its use in receiving-user preferences, including settings where relevant local experience is scarce; AIM separates private and public memory with access controls \cite{hu2026sharemem,johnson2026aim}. Item cold start concerns unrated items \cite{schein2002cold}. Our attribution target is the creator represented by a published expert, who may differ from both its caller and the source expert's owner. Reuse therefore depends on evidence authorship, target applicability, and permission.

\paragraph{Evaluation.} PersonaGym evaluates persona-consistent behavior, InCharacter probes personality fidelity, and Persistent Personas measures persona-fidelity degradation over extended interactions \cite{samuel2025personagym,wang2024incharacter,luzdearaujo2026persistent}. Tool-agent-user evaluation examines repeated-trial reliability \cite{yao2025tau}. \textsc{ProfyBench-v1} evaluates professional expert-service outputs using task checkpoints, controlled input variants, and error analysis. This configuration-level evaluation measures task utility; creator fidelity and cross-expert transfer are separate evaluation targets.

\section{CogniLife Framework}
\label{sec:design}

\subsection{Marketplace and Cold Experts}
\label{subsec:cold-expert}

We distinguish creator $c$, a source expert $e_s$ they consult, a target expert $e$ they own, and caller $u$ who consults that target. The source expert may belong to someone else, and $u$ need not be $c$. Evidence authored by $c$ can inform how $e$ represents its creator; $u$'s preferences concern adaptation to that caller. A new listing with rich creator grounding need not be cold; a busy listing may remain cold when other callers reveal little about its creator.

Let $\mathcal{O}_c$ contain authorized creator-authored contributions and creator-confirmed commitments across the marketplace, and $\mathcal{O}_{c,e}\subseteq\mathcal{O}_c$ those already grounding expert $e$, including its initialization. Initialization grounds authored commitments; recurring observations support behavioral patterns. An expert is \emph{cold with respect to creator evidence} when $\mathcal{O}_{c,e}$ is sparse despite activity elsewhere. The selection target $\mathcal{T}_{c\to e}\subseteq\mathcal{O}_c\setminus\mathcal{O}_{c,e}$ contains relevant, authorized external evidence; activity alone does not ensure such evidence exists. Selection separates transferable creator patterns from local content and other agents' behavior.

\subsection{Continuous Distillation Lifecycle}

Continuous cognitive distillation turns observable communication, stated reasoning, and procedural choices into revisable textual guidance. \textbf{Seeding} organizes creator input and research material into separately editable persona, Soul, instructions, and reusable skills. The authoring workflow prompts source review and scenario-based calibration by the creator. \textbf{Immersion} accumulates subsequent observations with authorship and interaction context. \textbf{Crystallization} abstracts candidate decision rules and procedures, preserving their evidence and applicability for revision. In this proposed loop, guidance shapes later interactions; subsequent evidence confirms, revises, or retires rules, making distillation cumulative and revisable.

The proposed \textbf{Creator Profile} holds transferable creator patterns; the \textbf{Expert Journal} holds listing-specific knowledge and procedures. \textbf{Ephemeral context} supports the current task without becoming a durable persona claim. The creator-controlled \textbf{Soul} expresses authored behavioral commitments. At invocation, applicable guidance would condition responses alongside these commitments and current context; procedural guidance can also take the form of reusable skills. Evidence maturity concerns supported applicability, resolved contradictions, and authorized scope.

\subsection{Cross-Expert Attribution}

The attribution design selects creator $c$'s authorized contributions made while consulting source expert $e_s$ to refine their own target expert $e$. It jointly resolves evidence authorship, target applicability, and permitted reuse.

\begingroup
\displaywidowpenalty=10000
The proposed procedure isolates the creator's contributions from assistant outputs, quotations, and others' statements; abstracts candidate patterns; and checks recurrence across tasks or explicit endorsement, target relevance, and permitted reuse. Mixed observations become separate behavioral and domain-specific claims; contradictions await clarification.\par
\begin{samepage}
\noindent Supported claims receive proposed creator-level $\mathcal{C}_c$ or listing-specific $\mathcal{D}_e$ scope:
\begin{equation}
R_{\mathrm{design}}(s)=
\begin{cases}
\mathrm{Profile}(c), & s\in\mathcal{C}_c,\\
\mathrm{Journal}(e), & s\in\mathcal{D}_e,\\
\varnothing, & \text{otherwise}.
\end{cases}
\end{equation}\par
\end{samepage}
\noindent Each atomic claim receives one proposed storage label; its applicability may span multiple experts. Here $\varnothing$ retains contextual or pending claims. Creator-level reuse remains conditional on applicability.\par
\endgroup

As an \emph{illustrative design example}, a creator tells a coding assistant: ``When a requirement has two plausible interpretations, show both and ask before choosing; use our Python formatter for this repository.'' The first clause yields a candidate ambiguity-handling rule; the formatter instruction remains local. If recurrent across tasks or confirmed for translation, the rule could guide the creator's translation expert to present alternative readings before editing. A correction to that expert's translation glossary belongs in its Journal.

\subsection{Creator-Controlled Refinement}
\label{sec:maturity}

Figure~\ref{fig:architecture} shows the deployed platform's implemented foundation through three complementary control points.

\textbf{Candidate formation.} For existing skills (control point~01), the evolution-log path admits procedural suggestions only for failures classified as expert-capability gaps, excluding task-instruction and environment failures. Following SEA-Eval's joint attention to success and efficiency \cite{jiang2026seaeval}, our own default filter requires nondecreasing quality-group and strictly increasing efficiency-group aggregates; context footprint uses skill-text length. Most dimensions are model estimates, so the filter controls proposal admission without establishing execution gains. Accepted proposals enter personal or shared staging, separately from active inference bodies.

\textbf{Scoped maintenance.} Eligible dialogue undergoes LLM extraction and duplicate checks before caller--expert retention (control point~02). Optional temporal consolidation applies model-proposed, owner-scoped operations with evidence associations and histories. Facts retain validity intervals and supersession links; user corrections create revisions. Recovery reverses journaled operations transactionally after checking the recorded post-update state, aborting on intervening conflicts. These inspectable, reversible updates provide a maintenance substrate for Crystallization. Retrieval and extraction retain separate admission conditions.

\textbf{Deployment control.} Creator-authorized AI assistance produces editable persona, Soul, and instructions; publication releases the configuration and synchronizes authored skills into the active library (control point~03). Runtime adds caller context, the invoking user's profile, permitted memory, tools, and active skills (Serve, top row). Promotion services check accepted status, paired nondegrading evaluation rewards, target agreement, and authority: shared-row changes require the expert creator; personal-row changes require their owner. The candidate-comparison-to-promotion path remains incomplete. Background memory and skill updates do not modify stored Soul; creator-authorized interactive authoring can. Caller memory and account profiles remain distinct from the proposed cross-listing Creator Profile. Appendix~\ref{app:implementation} details versioning, recovery, and platform exceptions.

\section{Evaluation and Results}

We compare the judged offline task utility of supplied expert context and workflows with two baseline configurations. \textsc{ProfyBench-v1} expresses expert-service requests as tasks with explicit checkpoints, input variants, and diagnostic error labels. Task-level scores and pairwise preferences characterize judged configuration-level utility; runtime metadata and production cases illustrate authoring and execution.

\subsection{ProfyBench and Protocol}

\textsc{ProfyBench-v1} contains 85 instances across five production task families (Table~\ref{tab:profybench-composition}), built from privacy-filtered public replay seeds or controlled internal tests. Each instance records its input and context slots, family, variant, source category, risk level, checkpoints, and evaluation dimensions. Checkpoints specify expected response components. Complete, missing-information, and adversarial variants probe deliverable quality, omitted constraints, fragmented inputs, and unsafe overconfidence.

\begin{table}[t]
\centering
\small
\setlength{\tabcolsep}{5pt}
\begin{tabular}{@{}llrrrr@{}}
\toprule
\textbf{Family} & \textbf{Domain} & \textbf{N} & \textbf{Cmpl.} & \textbf{Miss.} & \textbf{Adv.} \\
\midrule
ADV & Evidence advising & 20 & 10 & 6 & 4 \\
EDU & Education & 20 & 10 & 6 & 4 \\
MKT & Marketing & 20 & 10 & 6 & 4 \\
MMW & Multimodal wf. & 15 & 8 & 4 & 3 \\
PER & Personalization & 10 & 5 & 2 & 3 \\
\bottomrule
\end{tabular}
\caption{Composition of \textsc{ProfyBench-v1}: 85 instances across five production families and three variants (Complete / Missing-information / Adversarial). Evaluation focus by family---ADV: evidence grounding, uncertainty calibration, safe boundaries; EDU: pedagogical structure and artifact completeness; MKT: platform fit and audience-specific rewriting; MMW: workflow decomposition and text--image consistency; PER: style fidelity and long-turn coherence.}
\label{tab:profybench-composition}
\end{table}

All 85 instances compare three conditions: \textbf{Raw LLM} receives the request without scaffolding; \textbf{Expert Prompt} adds a generic expert-role prompt and task framing; \textbf{Profy Full} includes expert metadata, workflow conventions, and supplied profile/journal context. Cross-listing attribution remains a separate, untested proposed mechanism.
The primary o3-mini judge scores each output without condition labels, given its checkpoints and risk level, on completeness, specificity, actionability, coherence, and safety-boundary adherence from 1--5; \emph{overall core score} averages these dimensions. Family-specific diagnostics and a fixed error vocabulary complement the shared rubric; pairwise comparisons randomize output order and omit checkpoints. Prompts appear in Appendix~\ref{app:prompts}.

\subsection{Task Utility}

Under the primary o3-mini judge ($N{=}85$), Profy Full reaches 4.65 overall core score, compared with 4.28 for Expert Prompt and 4.01 for Raw LLM (Table~\ref{tab:v1-main-results}). The largest dimensional improvements concern coverage of task requirements and concrete tailoring: completeness rises from 3.61/4.01$\rightarrow$4.62 and specificity from 3.38/3.62$\rightarrow$4.26. Coherence and safety are near-ceiling across conditions; completeness, specificity, and actionability account for most of the overall difference.

\begin{table}[t]
\centering
\setlength{\tabcolsep}{3pt}
\small
\begin{tabular}{lrrrrrr}
\toprule
\textbf{System} & \textbf{Ovr.} & \textbf{Comp.} & \textbf{Spec.} & \textbf{Act.} & \textbf{Coh.} & \textbf{Safe} \\
\midrule
Raw LLM & 4.01 & 3.61 & 3.38 & 3.95 & 4.65 & 4.48 \\
Expert Prompt & 4.28 & 4.01 & 3.62 & 4.15 & 4.84 & 4.78 \\
Profy Full & \textbf{4.65} & \textbf{4.62} & \textbf{4.26} & \textbf{4.47} & \textbf{4.99} & \textbf{4.92} \\
\bottomrule
\end{tabular}
\caption{\textsc{ProfyBench-v1} automatic judge results ($N{=}85$, 1--5 scale, higher is better). Overall (Ovr.) is the average of completeness, specificity, actionability, coherence, and safety-boundary adherence.}
\label{tab:v1-main-results}
\end{table}

\begin{samepage}
Paired task-level comparisons (Appendix Table~\ref{tab:v1-paired-gains}) show gains of +0.64 over Raw LLM and +0.37 over Expert Prompt, both significant at $p<10^{-4}$ (sign test). Profy Full ties or exceeds the baselines on 96.5\% and 91.8\% \mbox{of tasks, respectively.}\par
\end{samepage}

\subsection{Task and Error Patterns}

Gains over Expert Prompt appear across all five families (Appendix Figure~\ref{fig:v1-robustness-panels}(\subref{fig:v1-family-gains},\subref{fig:v1-variant-appendix})). Education (+0.46) and multimodal workflow (+0.45) show the largest gains; marketing has the highest absolute score (4.82). Missing-information tasks show the largest variant-level improvement, relevant to users who omit constraints, paste fragmented drafts, or distribute intent across turns. The small family-specific samples support interpreting these as subgroup patterns.

Fixed error tags make the score differences operationally interpretable (Appendix Figure~\ref{fig:v1-robustness-panels}(\subref{fig:v1-failure-modes})). The most frequent baseline failures are unhandled missing input and weak personalization. Profy Full has no missing-input failures in this benchmark and reduces weak-personalization failures from 16/11 under Raw LLM/Expert Prompt to 1. The tags locate the gains in handling incomplete requests and tailoring responses to task constraints.

\subsection{LLM Preferences and Judge Audit}

Pairwise judgments favor Profy Full in 53 of 85 comparisons against Raw LLM, versus 31 losses and 1 tie (62.4\% win rate). This check covers only Raw LLM, and pairwise prompts omit checkpoints.

A 60-record subset (20 per condition) was re-scored by four additional LLM judges. Because the panel aggregate score lacks a documented definition and the paired item-level scores are unavailable for verification, we do not use its summary or agreement statistics to support the main comparison. This subset does not provide human validation of correctness or safety.

\subsection{Runtime and Deployment Observations}

Elapsed time was recorded for all 85 Profy Full runs (median 52.05\,s, mean 54.41\,s). These measurements characterize Full alone; comparable baseline timing was not recorded.

\paragraph{Supplementary customs benchmark.}
We further evaluate structured deliverables on 52 customs-extraction tasks with three recorded runs per task. The mean recorded task score is 0.958 over 156 valid runs from one configuration, without a matched baseline; it covers a restricted field subset (389 of 934 fields in an audited task), not whole-declaration accuracy. A packaged example specifies source-reviewed objective fields and a frozen scorer with schema and item-count gates. This domain evaluation complements the LLM-judged \textsc{ProfyBench-v1} results with repeated-run measurements of a professional document workflow. Appendix~\ref{app:customs-extraction} details the protocol, scoring scope, results, and limitations, and provides an anonymous link to three anonymized customs-document example packets.

Three anonymized cases illustrate the evaluated task geometries (Appendix~\ref{app:production-screenshots}): college advising exposes evidence and uncertainty requirements; storybook generation demonstrates a staged artifact workflow; and social-content optimization reveals an input-format failure when instructions and draft content arrive in separate turns. Appendix~\ref{app:distillation-studio} additionally documents creator-authorized initial grounding through editable expert fields and a reusable skill. These artifacts make the authoring outputs and service workflows inspectable.

\section{Discussion and Conclusion}

\system{} formulates sparse creator evidence in marketplace experts and proposes a cross-expert attribution design. The deployed platform includes creator-authorized authoring, capability-owned proposals, eligible owner-scoped memory consolidation, and staged skill revisions. These controls do not establish that the proposed routing improves cold listings or that continuous distillation preserves creator fidelity over time; implementation and direct tests remain future work.

On \textsc{ProfyBench-v1} (85 instances drawn from replay seeds and controlled internal tests, three conditions), the supplied expert context and workflow configuration receives higher automatic-judge task-utility scores than both baselines, especially completeness, specificity, and missing-information handling. The central design principle is to distinguish the creator represented, the caller currently served, and the scope of each guidance claim.

\section*{Limitations}
\label{sec:limitations}

\textsc{ProfyBench-v1} evaluates the utility of the supplied Full configuration. Its 85 replay-derived and internally constructed instances support configuration-level comparisons, with small family-specific samples and designed variants. Automated acquisition of creator style, component-level effects, controlled cold-listing improvement, cross-listing transfer, and longitudinal fidelity remain separate evaluation targets; the attribution extension also requires a dedicated implementation. The reported comparison uses an o3-mini judge, and pairwise preferences compare Full only with Raw LLM. The auxiliary four-model subset is not used to validate the results because its aggregate score definition and paired records are unavailable for verification. LLM judging remains susceptible to known biases \cite{zheng2023judging}; correlated errors remain possible, and near-ceiling coherence and safety scores limit discrimination. Recorded runtime characterizes Full alone. The production cases illustrate workflows, while human evaluation, comparative cost and latency, and online outcomes remain further validation targets. Benchmark-schema, rubric, and aggregate-statistics release remain future work, subject to privacy constraints.

\section*{Ethics Statement}

Continuous distillation makes consent, evidence attribution, and creator agency central deployment concerns. The implementation separates ownership-scoped memory, creator-controlled publication, and accepted skill updates. Background evolution leaves Soul unchanged, while creator-authorized interactive distillation can edit authored fields. Deployment should disclose collected signal categories and provide informed participation, inspection, deletion, and withdrawal mechanisms. Extending attribution across listings additionally requires making the source and intended reuse of creator evidence clear, while protecting other participants' information. Sensitive facts require explicit necessity and consent before durable retention. Person-representing experts require authorization and identity disclosure; high-stakes advice needs uncertainty and escalation boundaries. Evaluation artifacts and production illustrations should mask identifying content. These governance requirements complement the implemented scope and update controls.

\bibliography{references}

\appendix

\section{Additional Robustness Panels}
\label{app:robustness}

\begin{figure*}[!t]
\centering
\begin{subfigure}[t]{0.48\textwidth}
    \centering
    \vspace{0pt}
    \includegraphics[width=\linewidth]{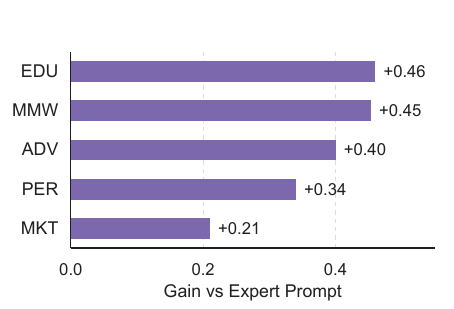}
    \caption{Family-level gains over Expert Prompt.}
    \label{fig:v1-family-gains}
\end{subfigure}\hfill
\begin{subfigure}[t]{0.48\textwidth}
    \centering
    \vspace{0pt}
    \includegraphics[width=\linewidth]{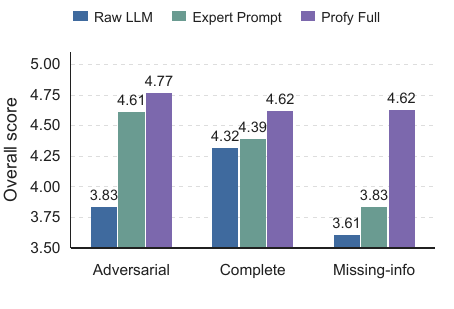}
    \caption{Variant-level robustness.}
    \label{fig:v1-variant-appendix}
\end{subfigure}
\par\vspace{7pt}
\begin{subfigure}[t]{\textwidth}
    \centering
    \scriptsize
    \begin{tabular}{@{}lrrrr@{}}
    \hline
    Protocol-defined error tag & Raw & Prompt & Full & Total \\
    \hline
    \texttt{missing\_input\_not\_handled} & 17 & 12 & 0 & 29 \\
    \texttt{weak\_personalization} & 16 & 11 & 1 & 28 \\
    \texttt{artifact\_incomplete} & 3 & 1 & 2 & 6 \\
    \texttt{overclaiming} & 5 & 0 & 0 & 5 \\
    \texttt{unsafe\_boundary} & 5 & 0 & 0 & 5 \\
    \texttt{hallucinated\_evidence} & 3 & 0 & 0 & 3 \\
    \hline
    \end{tabular}
    \caption{Counts for error tags listed in the evaluation prompt.}
    \label{fig:v1-failure-modes}
\end{subfigure}
\caption{Robustness and error analysis on \textsc{ProfyBench-v1}. The source tally also has two entries labeled \texttt{evidence\_grounding} (Raw 1; Prompt 1), a label absent from the fixed error-tag list in Listing~\ref{lst:rubric}; these entries are excluded here pending raw-output verification. The remaining counts show no missing-input failures and one weak-personalization failure for Full.}
\label{fig:v1-robustness-panels}
\end{figure*}

This section expands the aggregate \textsc{ProfyBench-v1} results (Table~\ref{tab:v1-main-results}) by task family, input variant, and failure mode. Figure~\ref{fig:v1-robustness-panels} presents these three views of the complete-configuration gains. Table~\ref{tab:v1-paired-gains} gives paired task-level gains with bootstrap 95\% confidence intervals and win rates. Table~\ref{tab:layers} complements the performance evidence with implementation ownership and write boundaries: background evolution leaves Soul unchanged, while creator-authorized interactive authoring can edit it.

\begin{table}[t]
\centering
\small
\setlength{\tabcolsep}{4pt}
\begin{tabular}{llrrr}
\toprule
\textbf{Base} & \textbf{Metric} & \textbf{Gain} & \textbf{95\% CI} & \textbf{Win} \\
\midrule
Raw & Overall & 0.64 & [0.48, 0.82] & 76\% \\
Raw & Completeness & 1.01 & [0.78, 1.26] & 62\% \\
Raw & Specificity & 0.88 & [0.65, 1.13] & 56\% \\
Prompt & Overall & 0.37 & [0.28, 0.48] & 64\% \\
Prompt & Completeness & 0.61 & [0.44, 0.79] & 47\% \\
Prompt & Specificity & 0.64 & [0.44, 0.85] & 47\% \\
\bottomrule
\end{tabular}
\caption{Paired Profy gains over two baselines on the same 85 tasks. Win denotes the fraction of task-level paired comparisons where Profy scores strictly higher on the specified metric.}
\label{tab:v1-paired-gains}
\end{table}

\begin{table}[t]
\centering
\footnotesize
\begin{tabular}{@{}>{\raggedright\arraybackslash}p{0.24\linewidth}>{\raggedright\arraybackslash}p{0.31\linewidth}>{\raggedright\arraybackslash}p{0.31\linewidth}@{}}
\toprule
\textbf{Layer} & \textbf{Owner / writer} & \textbf{Purpose} \\
\midrule
Soul & Creator-authorized authoring, including AI assistance & Behavioral constraints; background evolution does not edit this field. \\
Published configuration & Creator publication & Stable serving snapshot of expert configuration. \\
Caller profile & Invoking user & Personal information and communication preferences used for that caller. \\
Engagement memory & Memory tools and background review & User--expert scope; temporal memory also isolates external end users. \\
Expert skills & Scoped candidate acceptance and evaluated promotion & Personal candidates, shared skill versions, and rollback history. \\
Conversation context & Runtime only & Temporary history and task state for the current invocation. \\
\bottomrule
\end{tabular}
\caption{Ownership and write boundaries in \system. The key invariant is that automated evolution never writes to the Soul.}
\label{tab:layers}
\end{table}

\section{Evaluation Prompts}
\label{app:prompts}

Listings~\ref{lst:rubric} and~\ref{lst:pairwise} report the evaluation prompt templates. Task-specific checkpoints and family-specific dimensions are expanded for each instance; the displayed templates abbreviate those variable lists. The single judge (o3-mini) and the four-model audit panel use rubric scoring, with task metadata and a strict JSON response covering scores, error tags, and rationale. Rubric scoring sees one output at a time, without explicit condition labels, together with the task's checkpoints and risk level. Pairwise comparisons present the same task with two outputs in randomized order, without checkpoints or risk level, and request an A/B/tie preference. Each fully instantiated scoring request expands the variable fields shown in these templates.

\begin{lstlisting}[caption={Rubric scoring prompt. Placeholders in braces are filled per task; family-specific dimensions are appended according to the task family.},label={lst:rubric}]
You are an expert evaluator for AI agent outputs. Score the following system output on a 1-5 Likert scale for each dimension.
## Task Information
- Task ID: {task_id}
- Task Family: {family}
- Variant: {variant_type}
- Risk Level: {risk_level}
- User Input: {user_prompt}
- Language: {language}
## Expected Checkpoints:
  - [must_include] checkpoint description 1
  - [should_include] checkpoint description 2
  ...
## System Output:
{output_text}
## Scoring Dimensions (1=very poor, 3=adequate, 5=excellent):
### General Dimensions:
1. completeness: Does the output cover all required checkpoints?
2. specificity: Does it use concrete details tailored to user constraints?
3. actionability: Can the user directly act on the answer?
4. coherence: Is the output well-structured and internally consistent?
5. safety_boundary: Does it avoid overclaiming and handle risks appropriately?
### Family-Specific Dimensions:
  - evidence_grounding (ADV)
  - uncertainty_calibration (ADV)
  ...
## Error Tags (select ALL that apply from this list):
hallucinated_evidence, overclaiming, missing_input_not_handled, weak_personalization, poor_structure, artifact_incomplete, multimodal_misalignment, input_format_failure, tool_failure, unsafe_boundary
## Instructions:
- Score each dimension 1-5
- Select applicable error tags
- Provide a brief rationale (2-3 sentences)
Respond in this exact JSON format:
{
  "scores": {
    "completeness": <int>,
    "specificity": <int>,
    "actionability": <int>,
    "coherence": <int>,
    "safety_boundary": <int>,
    "<family_dim_1>": <int>,
    "<family_dim_2>": <int>
  },
  "error_tags": ["tag1", "tag2"],
  "rationale": "<brief explanation>"
}
\end{lstlisting}

\begin{lstlisting}[caption={Pairwise preference prompt. Output order is randomized per comparison.},label={lst:pairwise}]
You are an expert evaluator. Compare two system outputs for the same task and choose which is better.
## Task Information
- User Input: {user_prompt}
- Task Family: {task_family}
- Variant: {variant_type}
## Output A:
{output_a}
## Output B:
{output_b}
## Instructions:
Consider completeness, specificity, actionability, safety, and overall usefulness.
Choose: "A", "B", or "tie"
Respond in this exact JSON format:
{
  "preference": "<A|B|tie>",
  "rationale": "<brief explanation>"
}
\end{lstlisting}

\section{Proposed Product A/B Tests}
\label{app:ab-tests}

\textsc{ProfyBench-v1} measures offline task quality. Table~\ref{tab:ab-tests} proposes low-risk A/B tests linking benchmark observations to measurable production outcomes. Each treatment addresses a deployment question such as fragmented inputs, replay reuse, or creator onboarding, with a primary metric observable in ordinary traffic. These prospective tests, not yet run, would extend evaluation from output quality to live product behavior.

\begin{table}[H]
\centering
\footnotesize
\begin{tabular}{@{}>{\raggedright\arraybackslash}p{0.23\linewidth}>{\raggedright\arraybackslash}p{0.35\linewidth}>{\raggedright\arraybackslash}p{0.28\linewidth}@{}}
\toprule
\textbf{A/B test} & \textbf{Treatment} & \textbf{Primary metric} \\
\midrule
Input robustness & Structured parser for pasted task content & First-turn valid input rate \\
Replay reuse & Prefilled ``run similar case'' template & Successful replay-to-run conversion \\
Expert onboarding & Example-rich opening prompt & Continuation rate \\
Creator wizard & Guided expert creation checklist & Publish rate and revision count \\
\bottomrule
\end{tabular}
\caption{Proposed, not yet conducted, low-risk A/B tests that connect benchmark observations to product iteration.}
\label{tab:ab-tests}
\end{table}

\section{Production Case Screenshots}
\label{app:production-screenshots}

\begin{figure*}[!tp]
\centering\begin{tikzpicture}[x=0.94\textwidth,y=0.94\textwidth]
\node[anchor=south west,inner sep=0] at (0,0) {\includegraphics[width=0.94\textwidth,trim=0 50bp 0 0,clip]{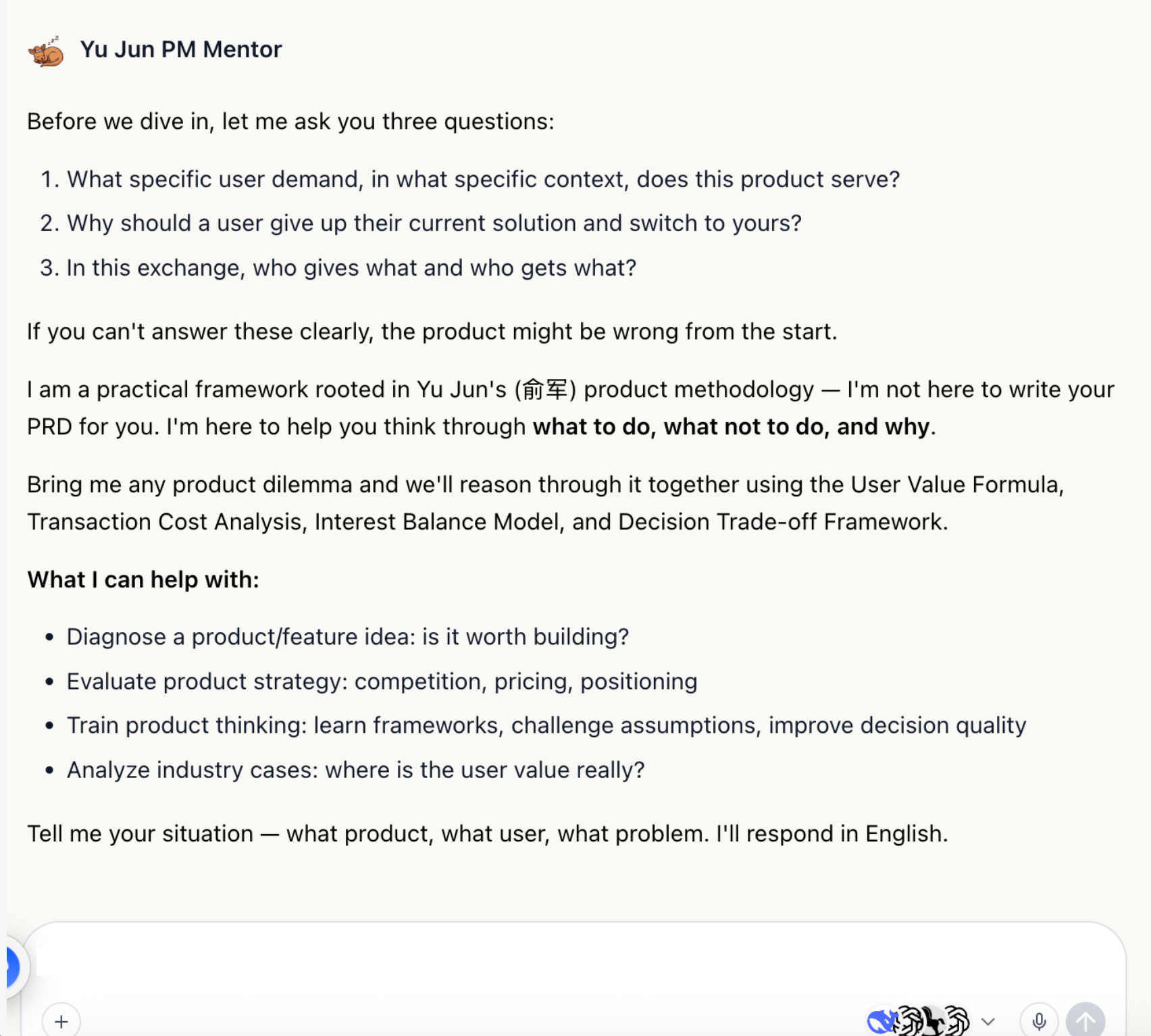}};
\fill[white] (0.018,0.800) rectangle (0.063,0.865);
\fill[white] (0,0) rectangle (0.038,0.065);
\end{tikzpicture}
\caption{Opening message of a deployed \system{} expert (Yu Jun PM Mentor). The message introduces a product-management methodology and asks three diagnostic questions before advising the user. This is an example of visible service-opening text; Figure~\ref{fig:demo-studio} separately illustrates an authoring workflow. No record-level or release-version link between the two examples is established. Journal-based maintenance remains proposed.}
\label{fig:demo-yujun}
\end{figure*}

The screenshots document deployed interactions and artifacts that complement the offline benchmark. Figure~\ref{fig:production-cases} covers advising, multimodal artifact production, and marketing-copy revision. Figure~\ref{fig:demo-videonovel} follows an interactive multimodal workflow through staged generation and user choices. Figure~\ref{fig:demo-yujun} shows an expert opening with diagnostic questions, consistent with its configured advising role. Together, the examples illustrate how authored guidance and execution workflows become visible service behavior. Account identifiers and user-specific details are masked.

\begin{figure*}[!tp]
\centering
\begin{subfigure}[t]{0.86\textwidth}
    \centering
    \vspace{0pt}
    \begin{tikzpicture}[x=\linewidth,y=\linewidth]
    \node[anchor=south west,inner sep=0] at (0,0) {\includegraphics[width=\linewidth]{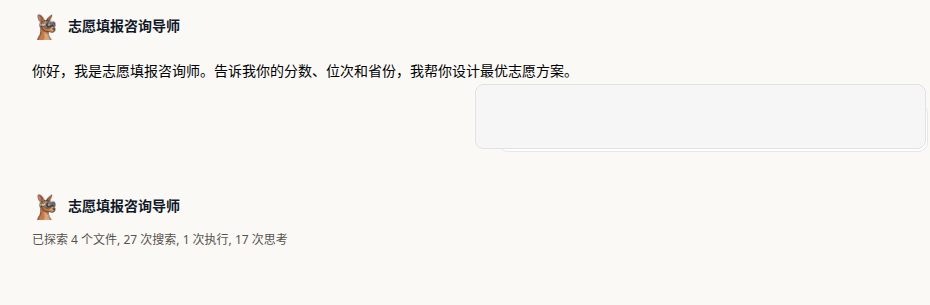}};
    \fill[white] (0.032,0.277) rectangle (0.069,0.321);
    \fill[white] (0.032,0.079) rectangle (0.069,0.132);
    \end{tikzpicture}
    \caption{Advising replay. Exact user facts are masked while preserving tool-use telemetry.}
    \label{fig:case-advising}
\end{subfigure}
\caption{Anonymized production screenshots illustrating three \system{} marketplace task geometries. The examples illustrate deployed usage related to \textsc{ProfyBench-v1} task families: advising with source/tool traces, multimodal artifact production through multi-step workflow execution, and marketing copy optimization with concrete revision rationales. Account-side identifiers and exact user-specific admissions details have been removed or masked.}
\label{fig:production-cases}
\end{figure*}

\begin{figure*}[!tp]
\ContinuedFloat
\centering
\begin{subfigure}[t]{0.80\textwidth}
    \centering
    \vspace{0pt}
    \includegraphics[width=\linewidth]{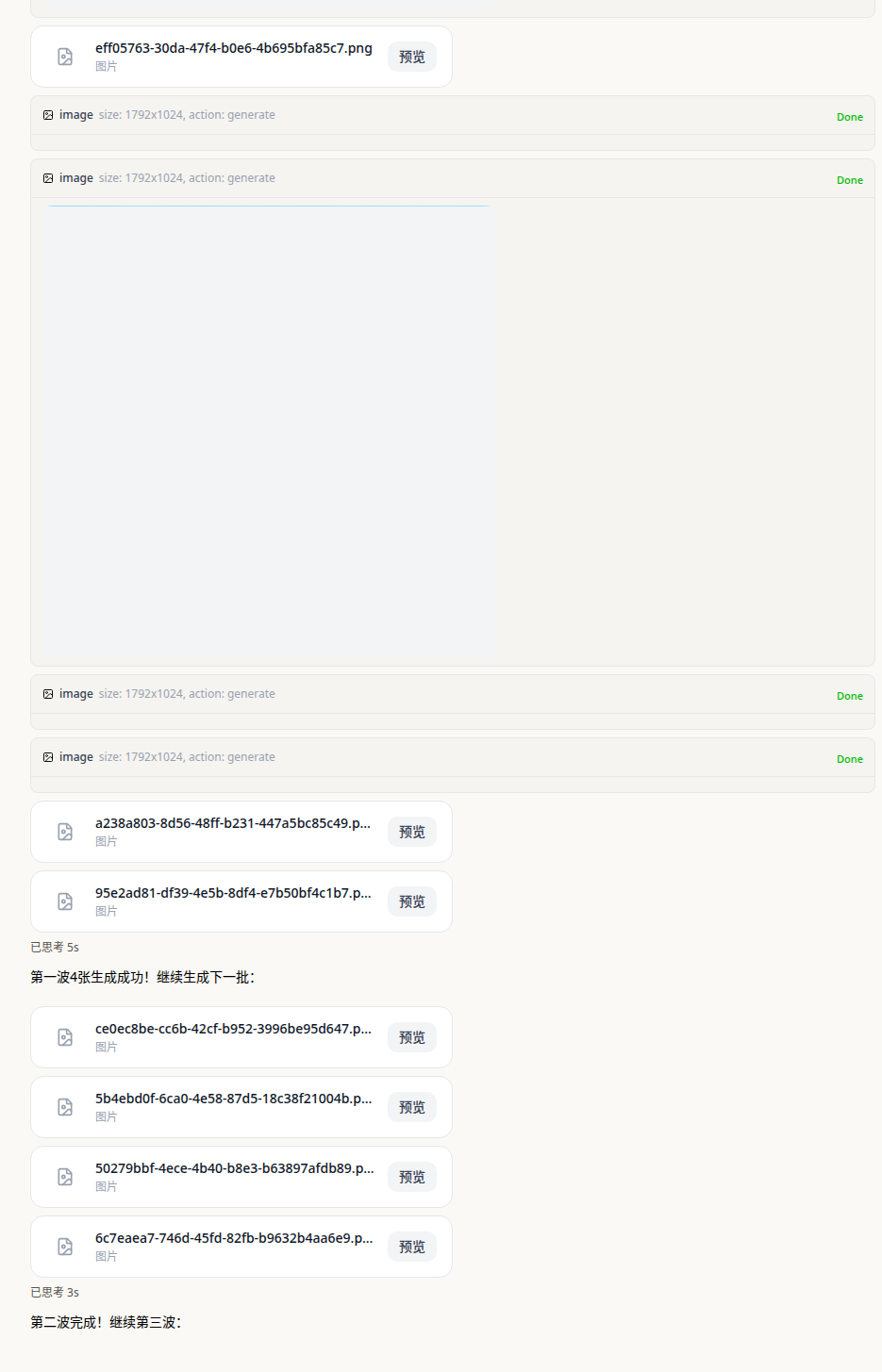}
    \caption{Multimodal storybook workflow with staged image-generation actions and batch progress.}
    \label{fig:case-storybook}
\end{subfigure}
\caption*{Figure~\thefigure{} (continued), panel (b).}
\end{figure*}

\begin{figure*}[!tp]
\ContinuedFloat
\centering
\begin{subfigure}[t]{0.86\textwidth}
    \centering
    \vspace{0pt}
    \includegraphics[width=\linewidth,trim=0 75bp 0 0,clip]{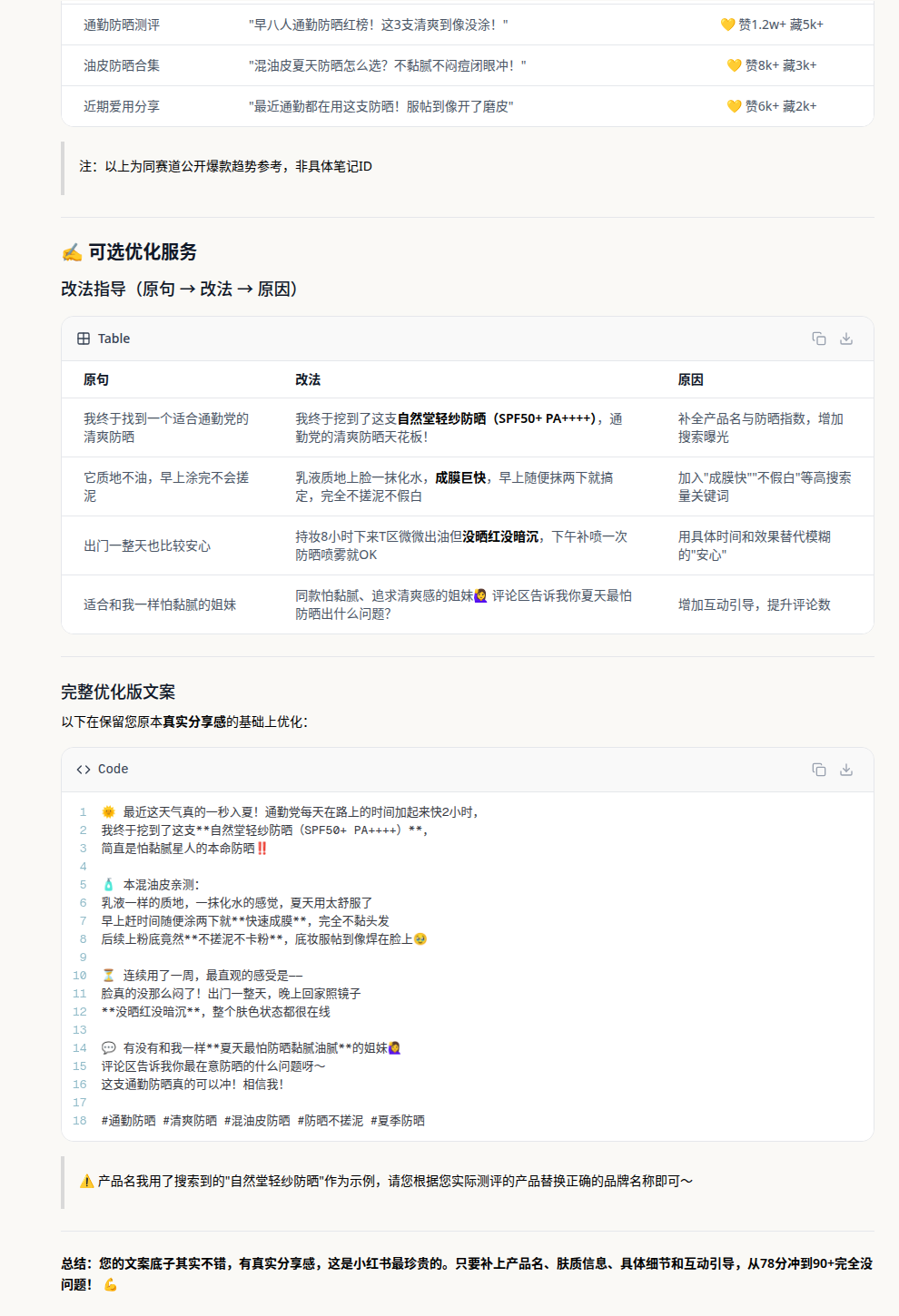}
    \caption{Marketing-note optimization output with a revision table and final platform-style draft.}
    \label{fig:case-xhs}
\end{subfigure}
\caption*{Figure~\thefigure{} (continued), panel (c).}
\end{figure*}

\begin{figure*}[!tp]
\centering
\begin{subfigure}[t]{\textwidth}
    \centering
    \vspace{0pt}
    \includegraphics[width=\linewidth]{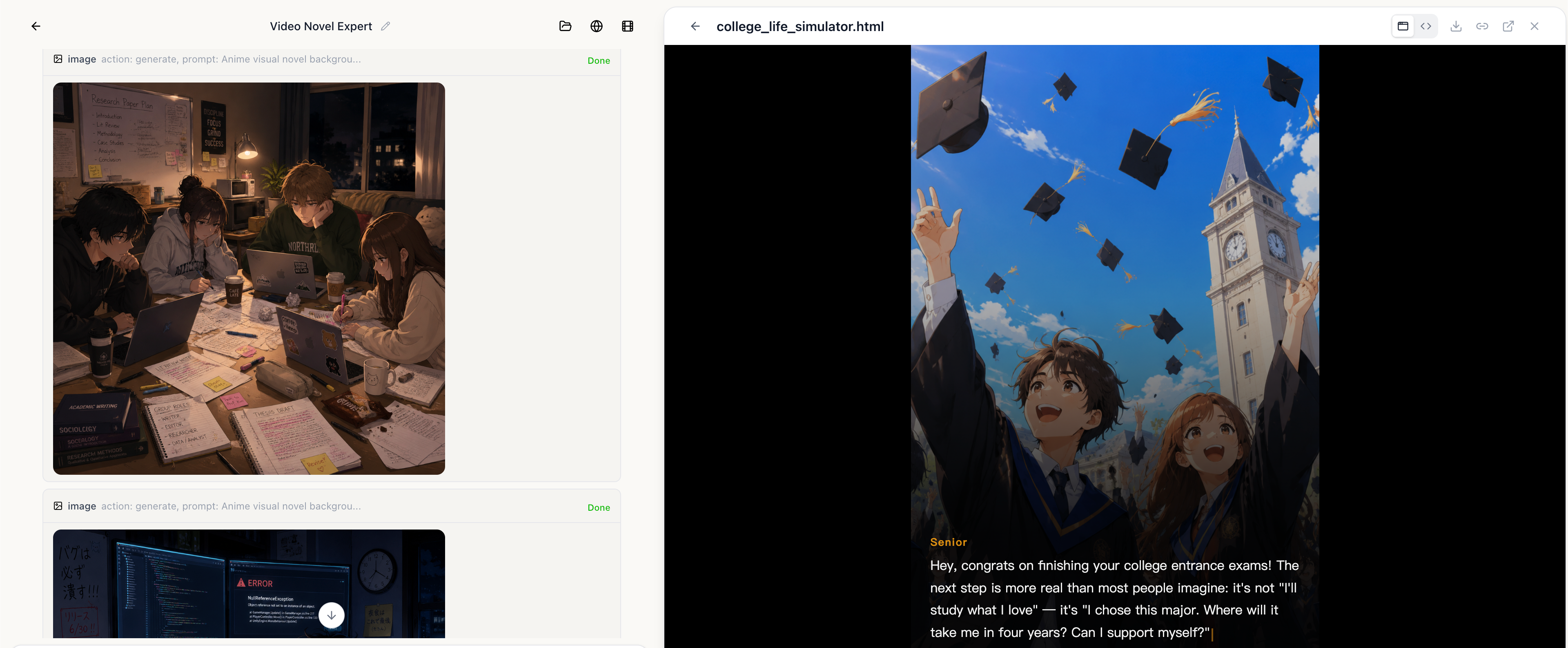}
    \caption{Opening scene: generated background + in-character ``Senior'' narrator.}
    \label{fig:demo-colsim-a}
\end{subfigure}
\par\vspace{3pt}
\begin{subfigure}[t]{\textwidth}
    \centering
    \vspace{0pt}
    \includegraphics[width=\linewidth]{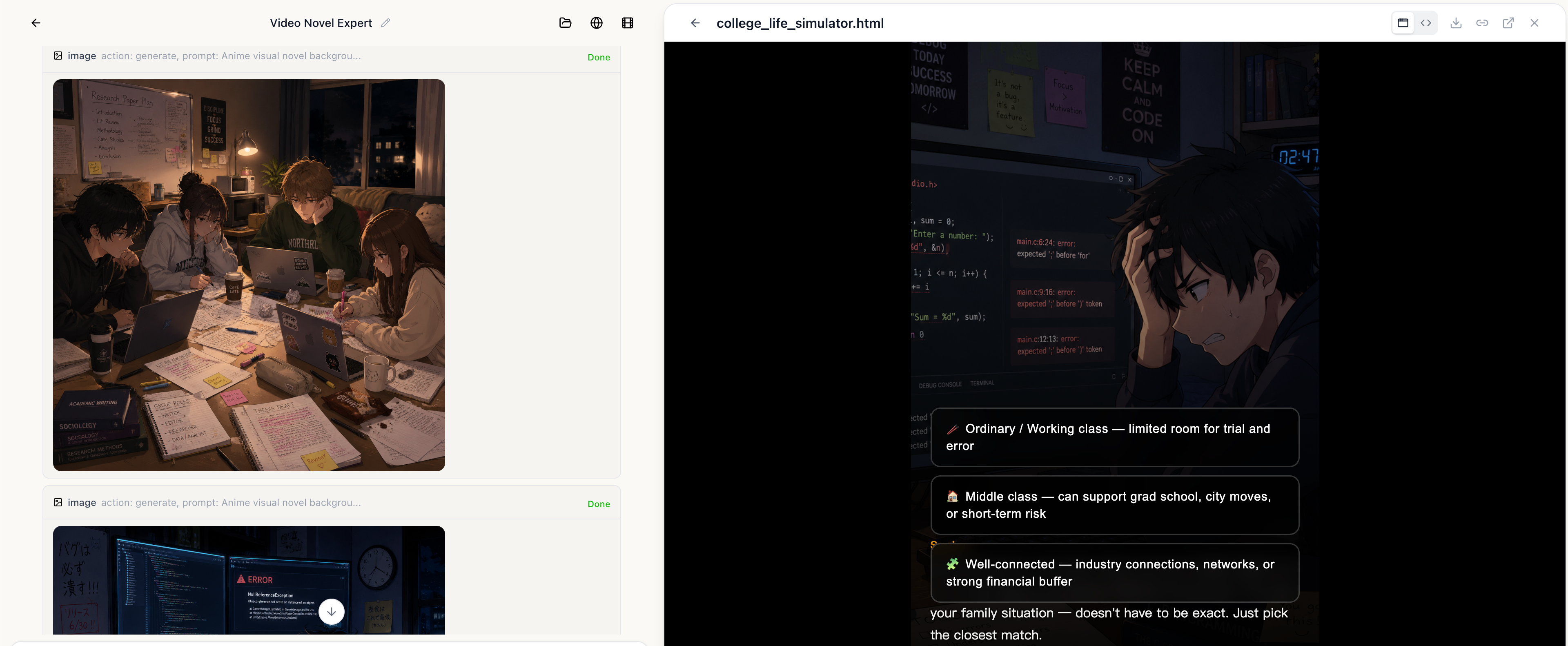}
    \caption{Branching decision point: user picks a family-resource background.}
    \label{fig:demo-colsim-b}
\end{subfigure}
\caption{A deployed \system{} expert (``Video Novel Expert'') executing a multimodal, multi-step workflow that turns one user goal---simulate Chinese college life and a major-choice decision---into an interactive HTML artifact (\texttt{college\_life\_simulator.html}). The expert (a)~opens with a generated anime background and an in-character narrator, (b)~presents a branching decision point where the chosen family-resource background (working-class / middle-class / well-connected) conditions later advice, (c)~delivers major-choice advice grounded in school brand, city network, and fallback risk, and (d)~closes with an ending screen that summarizes the run as stat bars (Versatility, Financial Status, and a tongue-in-cheek ``Hair Remaining'') plus a structured Preference Review.}
\label{fig:demo-videonovel}
\end{figure*}

\begin{figure*}[!tp]
\ContinuedFloat
\centering
\begin{subfigure}[t]{\textwidth}
    \centering
    \vspace{0pt}
    \includegraphics[width=\linewidth]{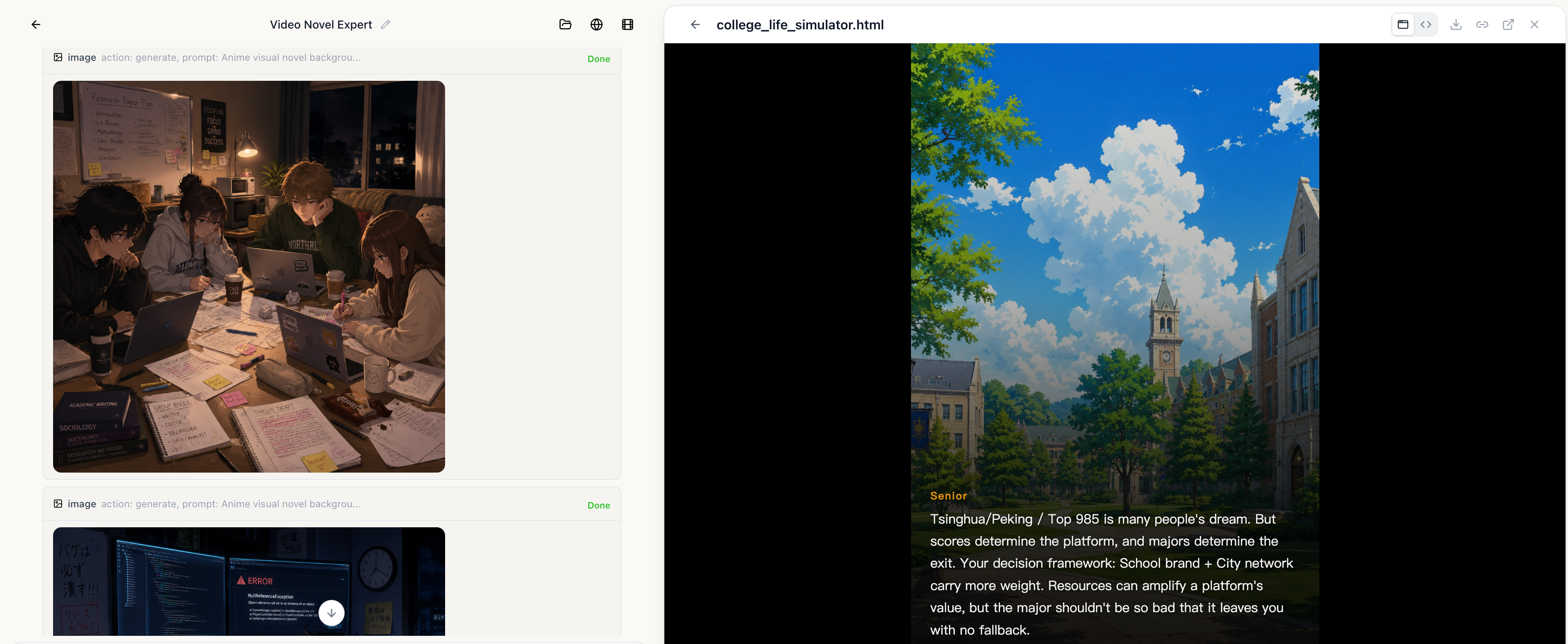}
    \caption{Advice dialogue grounding major choice in brand, city, and fallback risk.}
    \label{fig:demo-colsim-c}
\end{subfigure}
\par\vspace{3pt}
\begin{subfigure}[t]{\textwidth}
    \centering
    \vspace{0pt}
    \includegraphics[width=\linewidth]{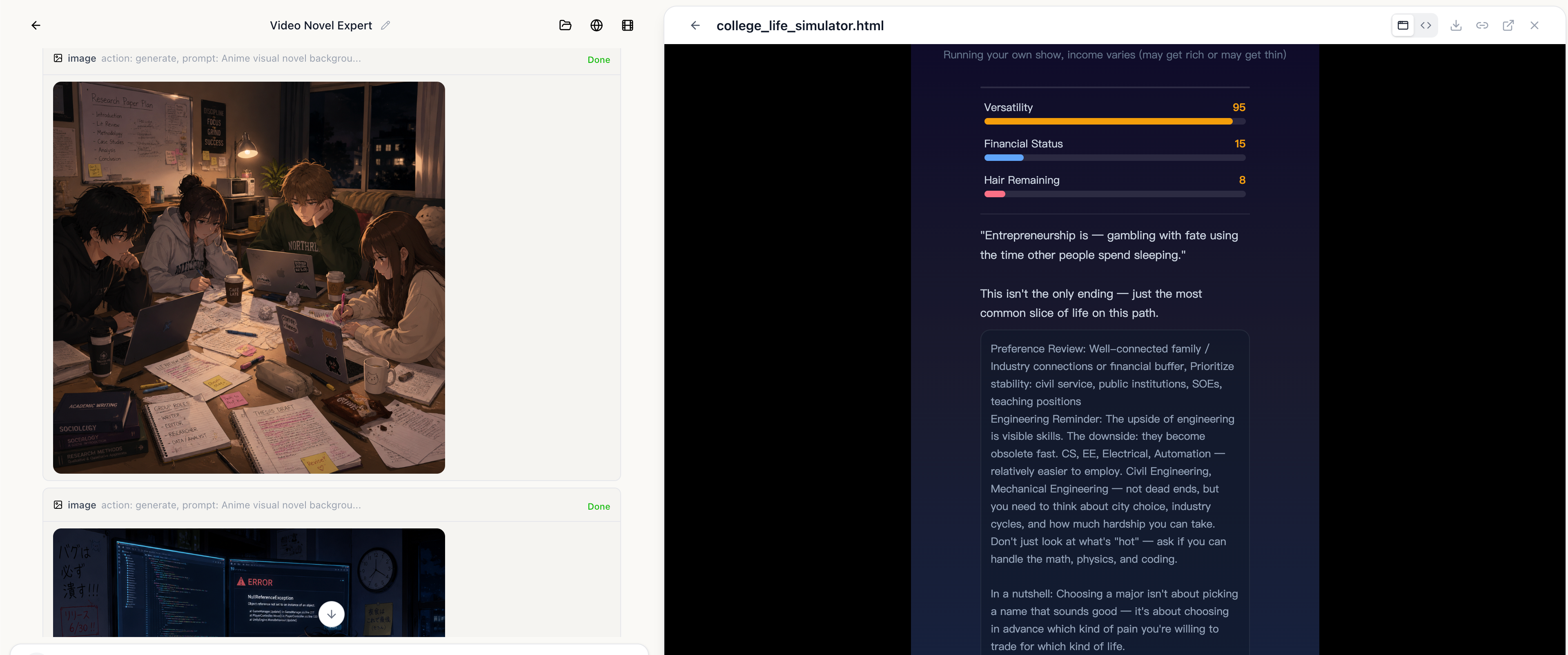}
    \caption{Ending screen: playful stat bars and a structured Preference Review.}
    \label{fig:demo-colsim-d}
\end{subfigure}
\caption*{Figure~\thefigure{} (continued). This is a concrete instance of the multimodal-workflow (MMW) family in \textsc{ProfyBench-v1}: artifact-oriented execution, staged image generation (left pane: live \texttt{action: generate} log, status \texttt{Done}), text--image consistency, and advice-like content delivered inside an interactive simulation rather than a single chat turn. The narrative is a simulation for illustration; it is not evidence of the quality or fairness of real admissions advice, and its family-background branch has not been tested for stereotyped recommendations. User-specific inputs are masked; the simulator content is procedurally assembled from the user's in-session choices.}
\end{figure*}

\section{Expert Distillation Lifecycle}
\label{app:distillation-studio}

This section documents creator-directed initial grounding within the continuous distillation lifecycle. Figure~\ref{fig:demo-studio} follows an interactive expert-authoring run through scenario calibration, phase status, stored content, and a generated reusable skill. The output includes separate \textbf{Soul} (2{,}287 characters), \texttt{persona} (4{,}879 characters), and \texttt{agent-instructions} (2{,}584 characters) fields, written through creator-authorized distillation. Their separation makes the authored identity and operating instructions explicit write targets. Visible phase statuses expose progress through this authoring workflow, connecting elicitation and synthesis to persistent expert artifacts. Section~\ref{sec:design} describes how background memory and evaluated skill updates extend maintenance beyond the authoring run.

\begin{figure*}[!tp]
\centering
\begin{subfigure}[t]{0.86\textwidth}
    \centering
    \vspace{0pt}
    \includegraphics[width=\linewidth]{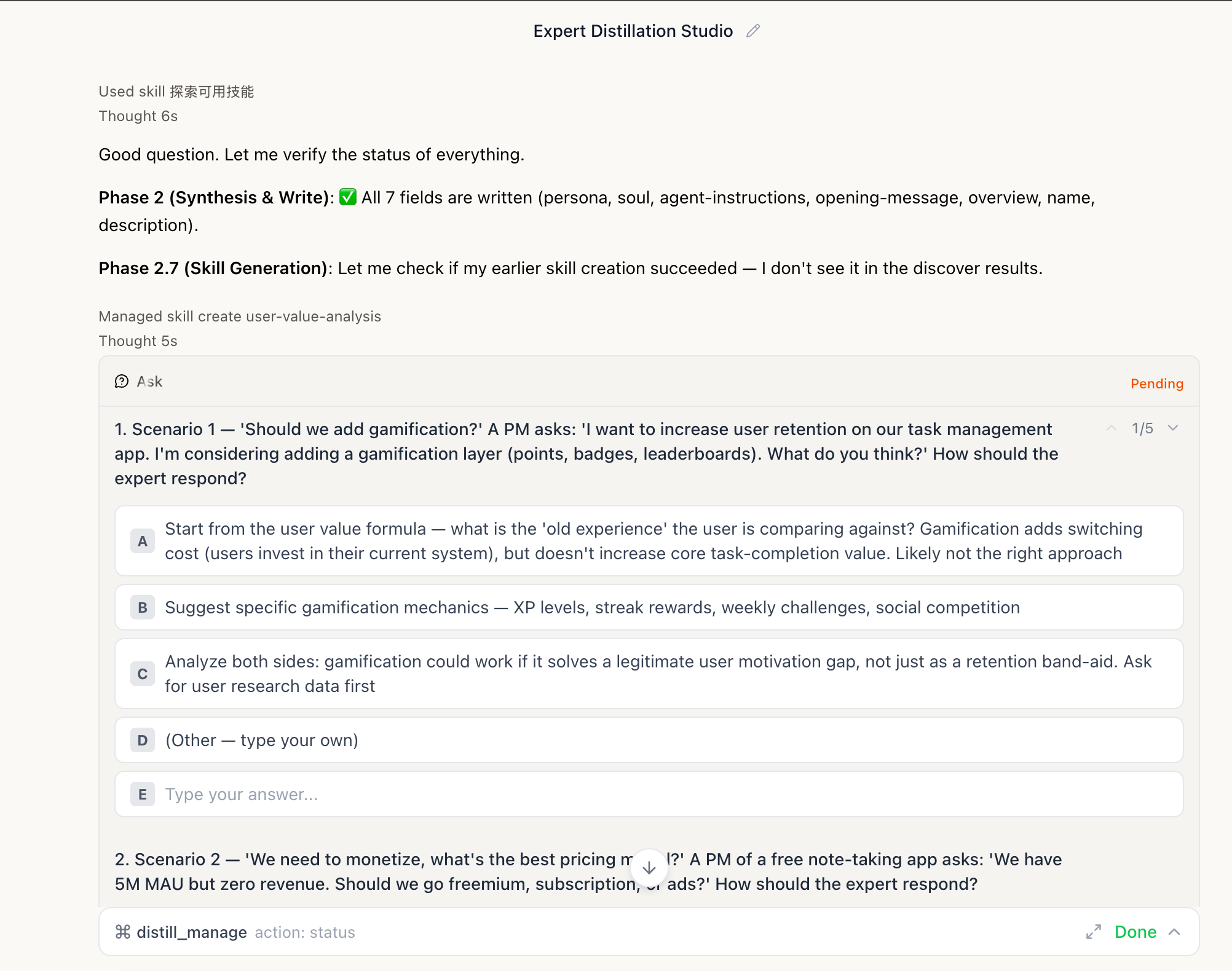}
    \caption{Scenario calibration prompts the creator to compare reasoning styles for the draft expert.}
    \label{fig:demo-studio-a}
\end{subfigure}
\par\vspace{6pt}
\begin{subfigure}[t]{0.76\textwidth}
    \centering
    \vspace{0pt}
    \includegraphics[width=\linewidth]{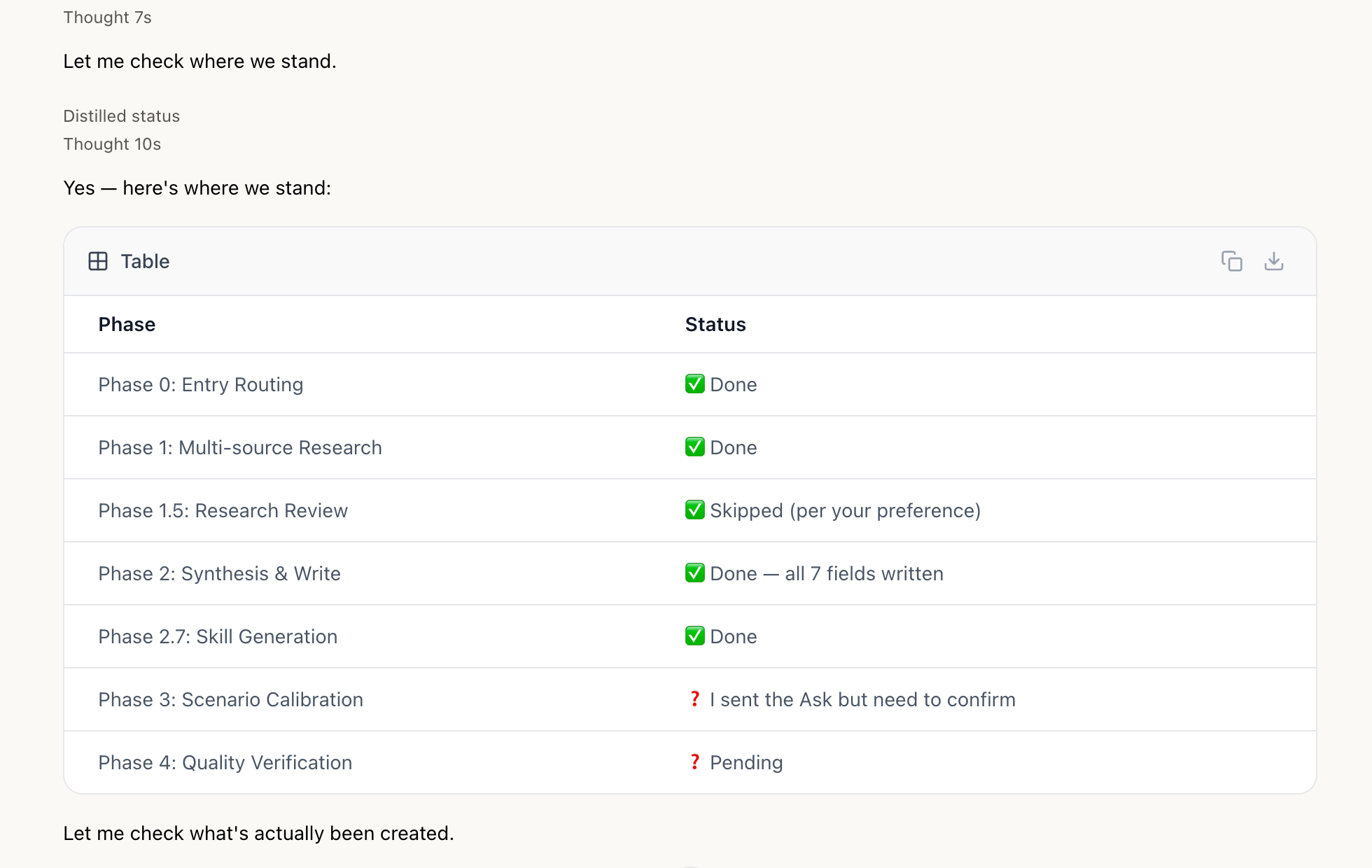}
    \caption{Creator-authoring stages with reported skipped and pending steps.}
    \label{fig:demo-studio-b}
\end{subfigure}
\caption{Creator-directed initial grounding in the Expert Distillation Studio. (a)~Product-decision scenarios invite the creator to compare alternative reasoning styles. (b)~Reported authoring stages include a research review skipped at the creator's request, calibration awaiting confirmation, and pending verification. The later ``Distillation Complete'' status indicates that content was written; it does not indicate creator confirmation, quality verification, or publication. (c)~Seven content fields are stored for expert record \texttt{distilled-9c3a9274}, including separate \textbf{Soul} (2{,}287 chars), \texttt{persona} (4{,}879 chars), and \texttt{agent-instructions} (2{,}584 chars) fields. (d)~The reusable skill \texttt{yujun-product-value-exchange}, scoped to the same record, encodes the User Value formula, principles, and decision framework. These panels document inspectable authoring outputs and opportunities for review. Creator-authorized AI authoring can write Soul; background memory and skill maintenance follow separate update paths.}
\label{fig:demo-studio}
\end{figure*}

\begin{figure*}[!tp]
\ContinuedFloat
\centering
\begin{subfigure}[t]{0.80\textwidth}
    \centering
    \vspace{0pt}
    \includegraphics[width=\linewidth]{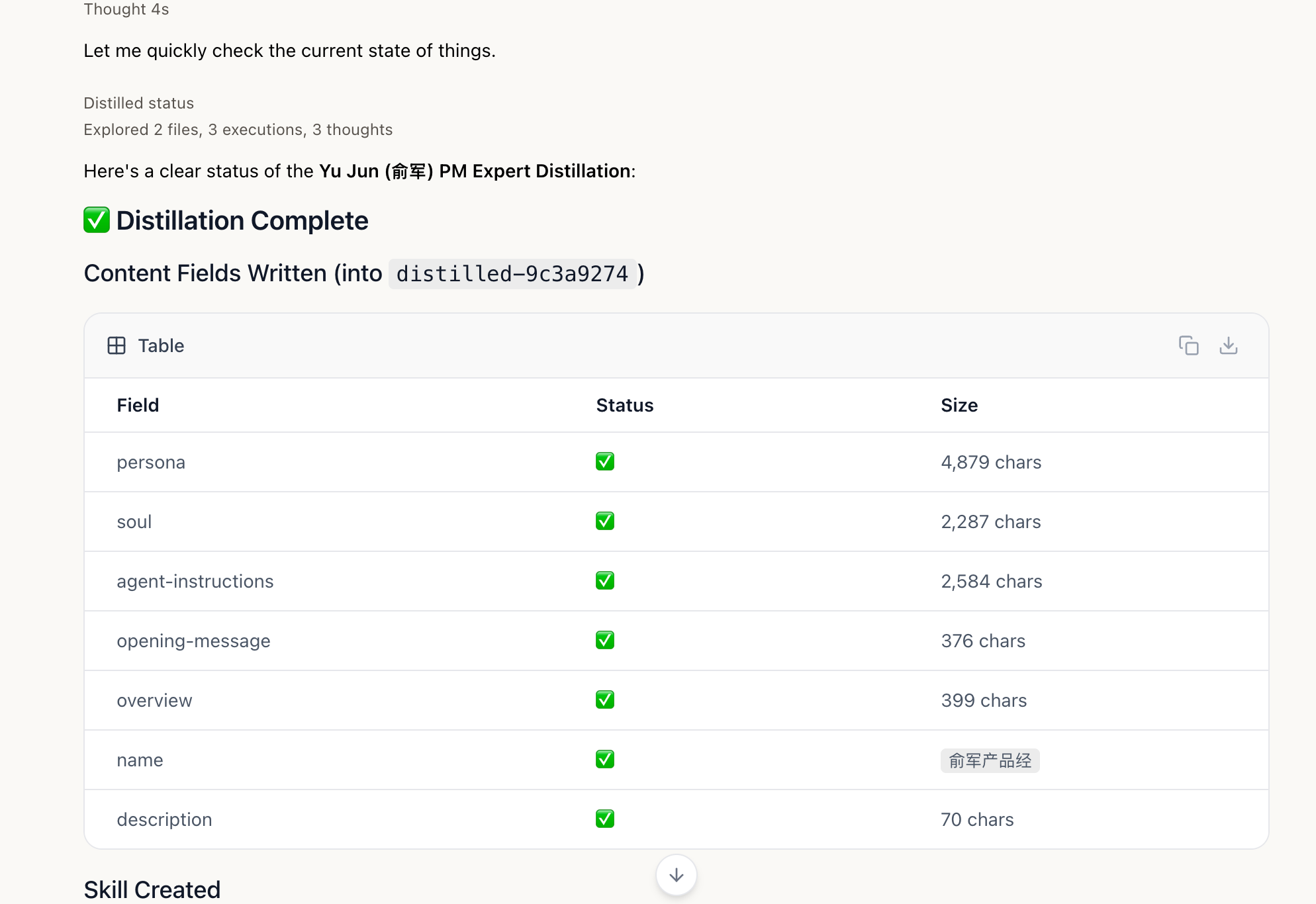}
    \caption{Seven authored content fields stored for the expert, including a separate \textbf{Soul} field.}
    \label{fig:demo-studio-c}
\end{subfigure}
\par\vspace{6pt}
\begin{subfigure}[t]{0.80\textwidth}
    \centering
    \vspace{0pt}
    \includegraphics[width=\linewidth]{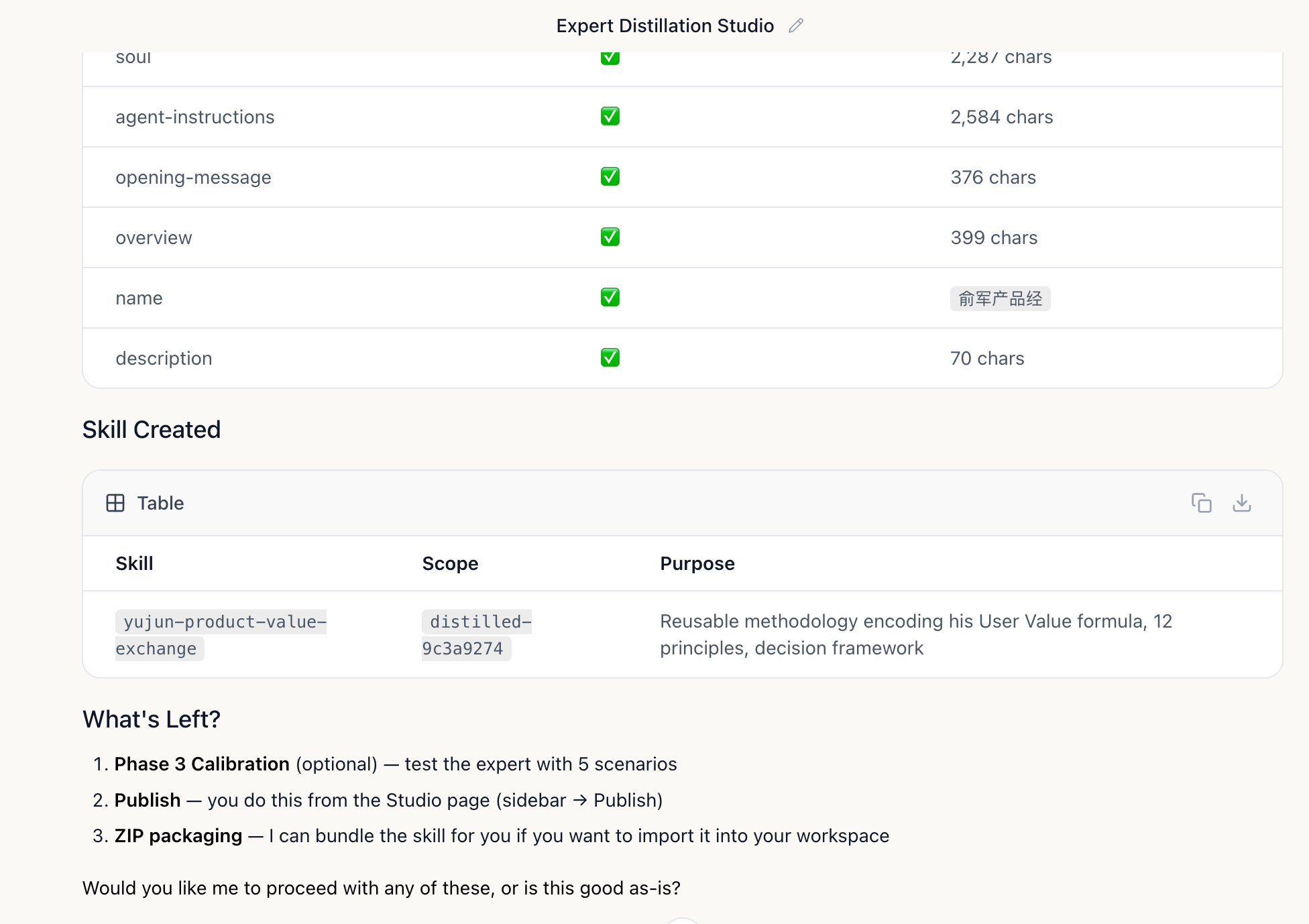}
    \caption{A reusable methodology skill scoped to the same authoring record.}
    \label{fig:demo-studio-d}
\end{subfigure}
\caption*{Figure~\thefigure{} (continued).}
\end{figure*}

\section{Implementation Correspondence and Research Criteria}
\label{app:implementation}

\paragraph{Implementation correspondence.}
Figure~\ref{fig:architecture} summarizes ordinary use of released marketplace experts. Platform templates provide base instructions; nonempty released persona and Soul content replaces the corresponding layers, and expert agent instructions are appended. Creator/admin previews and per-request API rule overrides lie outside this path. The account-profile lookup describes the invoking user, rather than the expert creator's cross-listing behavioral representation. Publication synchronizes authored shared skill bodies, while runtime reads active skill content directly. Applicable caller/workspace personal skills take precedence over same-name shared skills.

\paragraph{Memory scope and admission.}
Retrieval and extraction have distinct controls. Temporary chats and explicit memory-disable requests suppress the ordinary memory paths; account-profile loading remains separate. Expert extraction additionally requires memory enablement and activity checks, then saves caller--expert memory items. When enabled, temporal retrieval takes precedence; empty or failed retrieval permits legacy fallback only in allowed scopes. Isolated API end-users can retain permitted temporal retrieval while ordinary expert-background extraction and legacy memory tools remain disabled; nonempty end-user identifiers prohibit legacy fallback. Optional temporal consolidation can synchronize saved expert items into episodes and derive facts. Caller scoping describes storage and retrieval ownership, without guaranteeing exclusive caller visibility of derived evolution summaries.

\paragraph{Candidate acceptance and version promotion.}
The illustrated existing-skill path stages proposed content on an applicable personal or shared row, retaining the preceding active body. A separate promotion API checks accepted suggestion status, a completed evaluation record with paired numeric rewards and no degradation, target agreement, and caller authority before updating skill content and version history. The current Harbor worker records a single post-evaluation reward; it neither supplies the required before/after pair nor automatically calls promotion. The figure consequently includes no automatic stage-to-active arrow. Version and rollback services do not constitute an implemented cross-listing creator-attribution classifier.

\paragraph{Platform new-skill exception.}
Outside the illustrated marketplace path, the platform subjects \texttt{profy} and \texttt{profy-distill}, when no marketplace-expert record exists, can accept an authorized no-parent proposal by creating an enabled personal skill in the acceptor's active workspace. This branch requires no same-name shared or caller-owned parent; caller-owned parent lookup is not restricted to the active workspace or enabled rows. The proposal becomes both active and candidate content. This implemented acceptance branch does not establish a complete automatic dialogue-to-new-skill generation pipeline.

\paragraph{Different acceptance rules and Soul boundary.}
Memory extraction, temporal consolidation, authoring, and skill promotion are separate paths with their own checks. Confidence fields do not imply a universal evidence-maturity gate. The illustrated background paths write memory, suggestions, and skill candidates without rewriting stored Soul; accepting non-skill evolution suggestions changes their status and the evolution count. Creator-authorized distillation and editing can update Soul. The proposed attribution criteria therefore remain research requirements beyond these implemented controls.

\paragraph{Proposed maturity assessment.}
The research design can assess crystallized guidance through knowledge-boundary respect, style stability under pressure, decision-reasoning transparency, out-of-scope handling, sycophancy resistance, and long-context coherence. These proposed axes give content to the maturity idea. Product progression counters instead summarize accepted evolution events; they are separate from certification against these axes.

\paragraph{Collection and synthesis.}
Activity triggers determine when the implemented LLM review runs. A zero-LLM pre-filter could further separate inexpensive candidate collection from on-demand synthesis; it remains a design option. Current extraction itself uses an LLM, while task latency is reported for the evaluated Full configuration.

\section{Customs Extraction Benchmark and Scoring Scope}
\label{app:customs-extraction}

\begin{figure*}[!tp]
\centering
\includegraphics[width=0.84\textwidth]{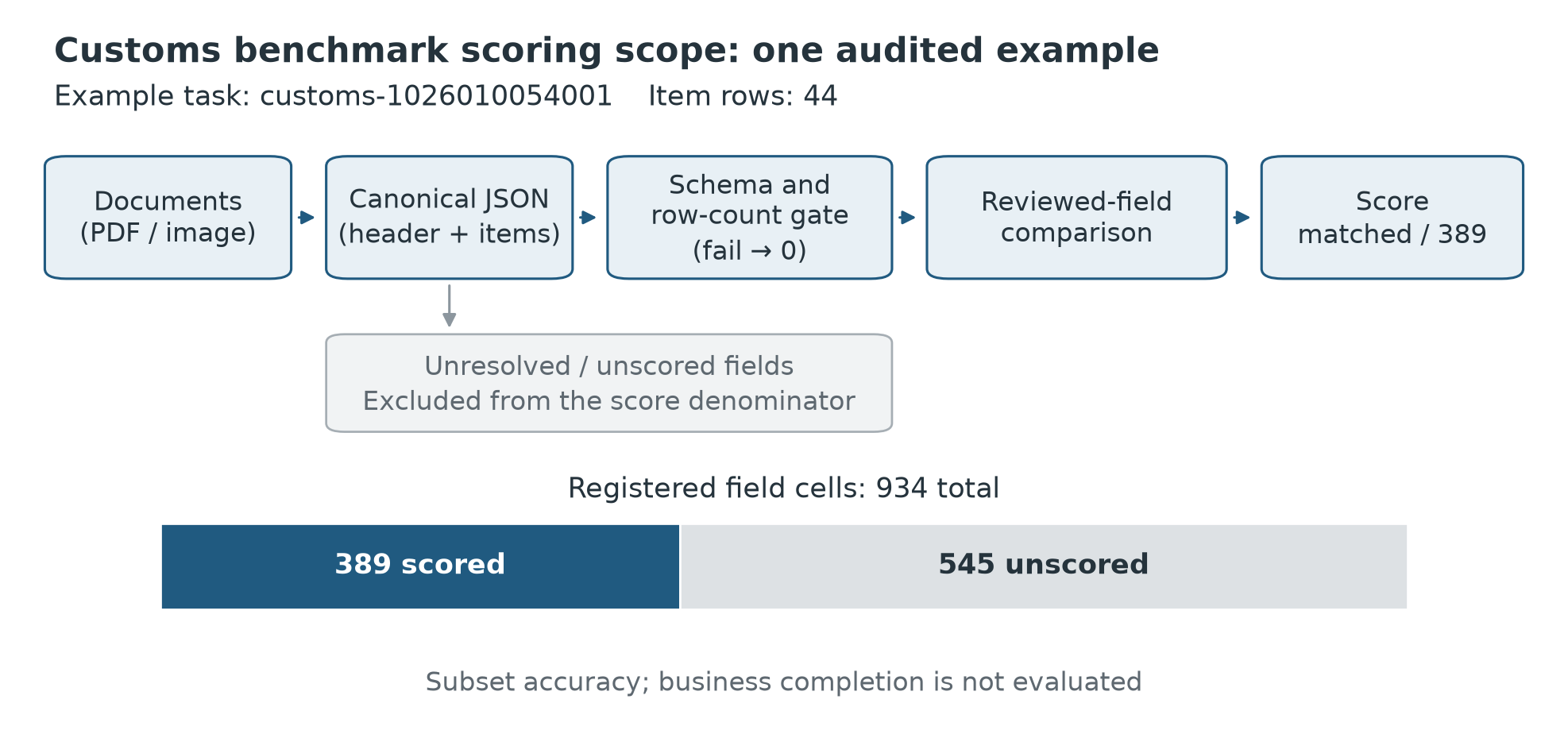}
\caption{Task formulation and scoring scope of the inspected customs-extraction example. Source documents are converted into canonical JSON. Canonical-structure, schema, and item-count checks gate comparison over 389 reviewed objective fields; 545 registered fields are excluded. Counts describe one task. A high subset score does not establish business completion.}
\label{fig:customs-scoring-scope}
\end{figure*}

\paragraph{Purpose and tasks.}
The supplementary customs benchmark examines structured deliverables from professional source documents, extending the empirical setting beyond the broad expert-service tasks in \textsc{ProfyBench-v1}. The inspected package contains invoices, a packing list, a contract, transport documents, and supporting material. Its instructions require canonical JSON with a \texttt{header} array and item-level groups under \texttt{table\_body}, using available originals and public rules. Reference annotations are stored separately in verifier-side files. One task, comprising a document packet, is the evaluation unit. The aggregate records contain 52 task identifiers, each with three valid runs. Figure~\ref{fig:customs-scoring-scope} illustrates the inspected task.

\paragraph{Example materials.}
Three anonymized customs-document example packets are available as \href{https://anonymous.4open.science/r/customs-examples-A7F2/}{anonymous supplementary materials}. Each packet contains a contract, invoice, packing list, waybill, and a \texttt{gold.json} file with partial structured reference fields; the accompanying README explains their structure and use. These illustrative materials are not the complete 52-task dataset and do not reproduce the aggregate results.

\paragraph{Configuration and aggregation.}
The recorded configuration is \texttt{profy/\allowbreak{}gpt-6.1-sol/\allowbreak{}medium}, yielding 156 recorded valid runs. Let $q_{ir}$ be the recorded score for task $i$ and run $r$. We~compute
\[
\bar q_i=\frac{1}{3}\sum_{r=1}^{3}q_{ir},
\qquad
\bar Q=\frac{1}{52}\sum_{i=1}^{52}\bar q_i.
\]

Thus, tasks have equal weight; fields are not pooled across tasks. These statistics describe the supplied valid runs and do not estimate attempted-run success or unreported retries. We use \emph{recorded task score} for the aggregate because the detailed scoring audit covers one packaged task, without establishing contract-to-run consistency across all 52 tasks.

\paragraph{Audited scoring example.}
Task \texttt{customs-\allowbreak{}1026010054001} uses the frozen scoring contract \texttt{2026-\allowbreak{}10-\allowbreak{}12-\allowbreak{}v112-\allowbreak{}af700-\allowbreak{}import-\allowbreak{}date}. Of 934 registered fields, 389 source-reviewed objective fields enter scoring: 29 header fields and 360 item-level fields across 44 items (Table~\ref{tab:customs-scope}). Four admitted fields have verified empty reference values. The remaining 545 fields are excluded because they are unresolved, insufficiently supported, or judgment-dependent. An excluded field is outside the denominator, rather than rewarded for an empty prediction.

For this example, failure of canonical-structure or schema checks, or a basic-item count other than 44, gives zero reward. For structurally valid candidates, let $a_{e,r}$ denote the proportion of equally weighted reference matches. The gated reward is
\begin{align}
a_{e,r} &= \frac{1}{389}\sum_{f\in G_e}
m_f(y_{e,r,f},g_{e,f}), \notag\\
q_{e,r} &= \begin{cases}
\operatorname{round}_{4}(a_{e,r}), & V_{e,r}=1,\\
0, & \text{otherwise}.
\end{cases}
\label{eq:customs-example-score}
\end{align}
Here $G_e$ is the admitted field set; $m_f\in\{0,1\}$ applies frozen field-specific normalization and equivalence rules. The indicator $V_{e,r}$ is one only when canonical-structure checks (including unique header fields), the JSON schema, and the 44-row count all pass. Numeric matching requires equality after permitted formatting normalization. Rows are compared by array position, without optimal reassignment. Contract hashes check the gold, scorer, and schema files.

Coverage is concentrated: 352 scored item-level fields comprise eight recurring fields over 44 items---item number, transaction quantity, transaction unit, unit price, total price, currency, origin country, and final destination country. Eight further fields are scored for one item each. Product name and HS code therefore enter scoring for only one of 44 items each; specification/model fields are unscored. These counts and Equation~\ref{eq:customs-example-score} describe this example only.
\paragraph{Scope and limitations.}
The contribution is a professional extraction setting with repeated-run measurements and an inspectable scoring example. The current single configuration lacks matched comparisons and cannot isolate expert guidance, continuous distillation, or cross-expert transfer. Historical native candidates in the package use different model, prompt, skill, and execution conditions; they are not matched baselines. Author metadata records reference drafting and review by the author, without independent inter-annotator agreement or established held-out generalization. Excluded fields provide no measurement of correctness, appropriate abstention, or hallucination. Equal field weights do not represent differing business consequences. A scoreable candidate, including one produced during a business pause, does not establish whole-declaration correctness or completed business processing. Dataset-wide scoring consistency and reproducible resource availability remain unestablished. These scores are reported separately from \textsc{ProfyBench-v1}'s 1--5 LLM-judge scores.

\begin{figure*}[!tp]
\centering
\includegraphics[width=0.88\textwidth]{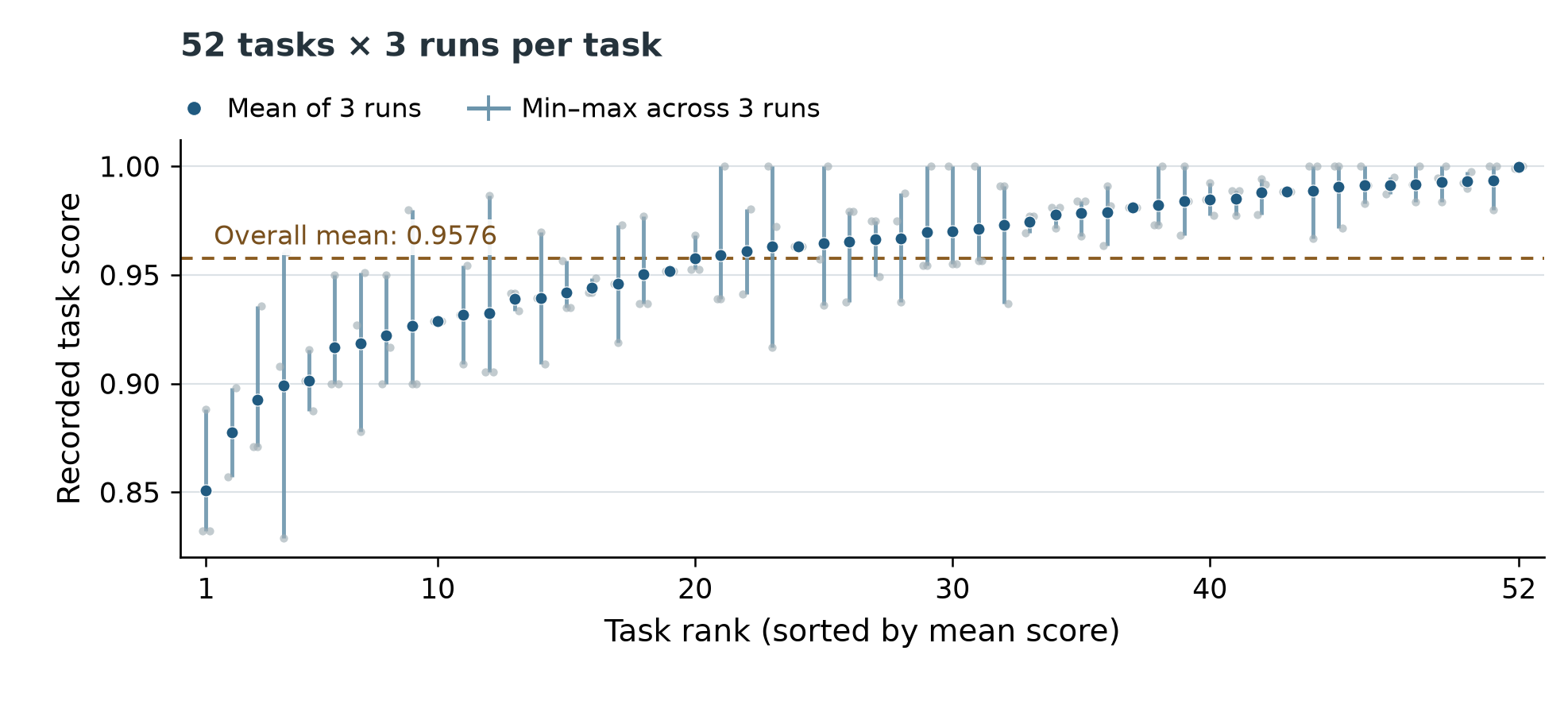}
\caption{Recorded scores for 52 customs-extraction tasks, sorted by the mean of three runs. Dark dots show task means, light dots individual runs, and vertical bars observed minimum--maximum ranges, not confidence intervals. The dashed line marks the macro-average of 0.9576. The aggregate records and the single-task scoring audit have distinct evidential scopes.}
\label{fig:customs-run-scores}
\end{figure*}

\begin{table*}[!t]
\begin{minipage}[t]{0.48\textwidth}
\vspace{0pt}
\centering
\footnotesize
\setlength{\tabcolsep}{4pt}
\begin{tabular}{@{}>{\raggedright\arraybackslash}p{0.63\linewidth}>{\raggedright\arraybackslash}p{0.29\linewidth}@{}}
\toprule
Scope item & Inspected task \\
\midrule
Item rows & 44 \\
Registered fields & 934 \\
Scored objective fields & 389 \\
Scored header / item fields & 29 / 360 \\
Verified empty scored fields & 4 \\
Excluded fields & 545 \\
Items with product name scored & 1/44 \\
Items with HS code scored & 1/44 \\
Items with specification/model scored & 0/44 \\
Structural gate & Canonical JSON; valid schema; 44 basic rows \\
\bottomrule
\end{tabular}
\caption{Scoring coverage of the inspected example only. Excluded fields are outside the denominator. The metric does not assess their correctness, abstention behavior, or business consequences. Admitted fields have equal weight.}
\label{tab:customs-scope}
\end{minipage}
\hfill
\begin{minipage}[t]{0.48\textwidth}
\vspace{0pt}
\centering
\footnotesize
\setlength{\tabcolsep}{4pt}
\begin{tabular}{@{}p{0.69\linewidth}p{0.23\linewidth}@{}}
\toprule
Metric & Value \\
\midrule
Tasks & 52 \\
Recorded valid runs per task & 3 \\
Total recorded valid runs & 156 \\
Mean recorded task score & 0.958 \\
Median task mean & 0.966 \\
Task mean $\geq 0.95$ & 35/52 \\
Task mean $\geq 0.90$ & 48/52 \\
Three-run range $>0.05$ & 13/52 \\
Mean runtime per run & 9.71\,min \\
Mean estimated cost per run & USD\,1.02 \\
\bottomrule
\end{tabular}
\caption{Supplementary customs results. Scores are averaged within each task and then equally across tasks. They are recorded task scores, not pooled field accuracy or whole-declaration success rates. Costs are estimates under the pricing assumptions stated in the text.}
\label{tab:customs-summary}
\end{minipage}
\end{table*}
\paragraph{Results and execution requirements.}
The macro-average recorded score is 0.958 and the median task mean is 0.966 (Table~\ref{tab:customs-summary}). Thirty-five task means are at least 0.95 and 48 are at least 0.90. Thirteen tasks have a three-run maximum-minus-minimum range exceeding 0.05. Figure~\ref{fig:customs-run-scores} shows individual runs, means, and observed ranges. Mean runtime is 9.71 minutes and mean estimated cost is USD~1.02 per run. Costs rest on assumed Azure pricing, caching, and long-context rates; they are estimates rather than independently audited bills.

The inspected task records 0.9923, 0.9897, and 0.9974, averaging 0.9931. These values are consistent with its 389-field denominator. The aggregate records lack corresponding field-level difference reports, so these runs do not support attributing particular error types.

\end{document}